\documentclass[10 pt, conference]{IEEEtran}
\IEEEoverridecommandlockouts

\usepackage{fancyhdr}

\usepackage[noadjust]{cite}

\usepackage{amsmath}
\usepackage{graphicx}
\usepackage{tabularx}
\usepackage{textcomp}
\usepackage{xcolor}
\usepackage{balance}
\usepackage{comment}
\def\BibTeX{{\rm B\kern-.05em{\sc i\kern-.025em b}\kern-.08em
    T\kern-.1667em\lower.7ex\hbox{E}\kern-.125emX}}
\usepackage[english]{babel}
\usepackage[utf8]{inputenc}
\usepackage{algorithm}
\usepackage[noend]{algpseudocode}
\usepackage{subcaption}
\usepackage[font = small]{caption}
\usepackage{multirow}
\usepackage{graphics}
\usepackage{adjustbox}
\usepackage{soul}

\usepackage{booktabs}
\usepackage{graphicx}

\begin{document}

\title{Memory Faults in Activation-sparse Quantized Deep Neural Networks: Analysis and Mitigation using Sharpness-aware Training}

\makeatletter
\newcommand{\linebreakand}{%
  \end{@IEEEauthorhalign}
  \hfill\mbox{}\par
  \mbox{}\hfill\begin{@IEEEauthorhalign}
}
\makeatother

\author{\IEEEauthorblockN{Akul Malhotra and Sumeet Kumar Gupta}
\IEEEauthorblockA{
Purdue University, West Lafayette, Indiana}


\thanks {{This work is supported by the Center for Brain-Inspired Computing (C-BRIC) and Center for Co-Design of Cognitive Systems (CoCoSys), funded by SRC and DARPA. We would also like to acknowledge Prof. Anand Raghunathan (Purdue University) and Dr. Sarada Krithivasan (IBM) for their suggestions. \textit{Corresponding Author}: \textit{Akul Malhotra}.} Email: \textit{malhot23@purdue.edu}.
}

}







\maketitle

\pagestyle{plain}

\begin{abstract}
 Improving the hardware efficiency of deep neural network (DNN) accelerators with techniques such as quantization and sparsity enhancement have shown an immense promise. However, their inference accuracy in non-ideal real-world settings (such as in the presence of hardware faults) is yet to be systematically analyzed. In this work, we investigate the impact of memory faults on \textit{activation-sparse} quantized DNNs (AS QDNNs). We show that a high level of activation sparsity comes at the cost of larger vulnerability to faults, with AS QDNNs exhibiting up to 11.13\% lower accuracy than the standard QDNNs. We establish that the degraded accuracy correlates with a sharper minima in the loss landscape for AS QDNNs, which makes them more sensitive to perturbations in the weight values due to faults. Based on this observation, we employ sharpness-aware quantization (SAQ) training to mitigate the impact of memory faults. The AS and standard QDNNs trained with SAQ have up to 19.50\% and 15.82\% higher inference accuracy, respectively compared to their conventionally trained equivalents. Moreover, we show that SAQ-trained AS QDNNs show higher accuracy in faulty settings than standard QDNNs trained conventionally. Thus, sharpness-aware training can be instrumental in achieving sparsity-related latency benefits without compromising on fault tolerance. 
\end{abstract}


\vspace{-0.1in}

\section{Introduction}
\label{sec:intro}

The remarkable success of deep neural networks (DNNs) for decision-making and sensory processing tasks has prompted the exploration of DNN edge accelerators \cite{accelerator}. However, the performance benefits of state-of-the-art DNNs come at the cost of large storage and computation requirements, introducing various design challenges, especially for energy- and memory-constrained edge applications. The need to reduce the size and computational complexity of DNNs has led to the emergence of Quantized Deep Neural networks (QDNNs). QDNNs utilize reduced bit precisions for their weights and activations to achieve energy and latency savings associated with storage, computation and communication, while maintaining an accuracy comparable to their full-precision counterparts \cite{google_quant}.  

\begin{figure}[t!]
\centering
  \includegraphics[width = 0.95\linewidth]{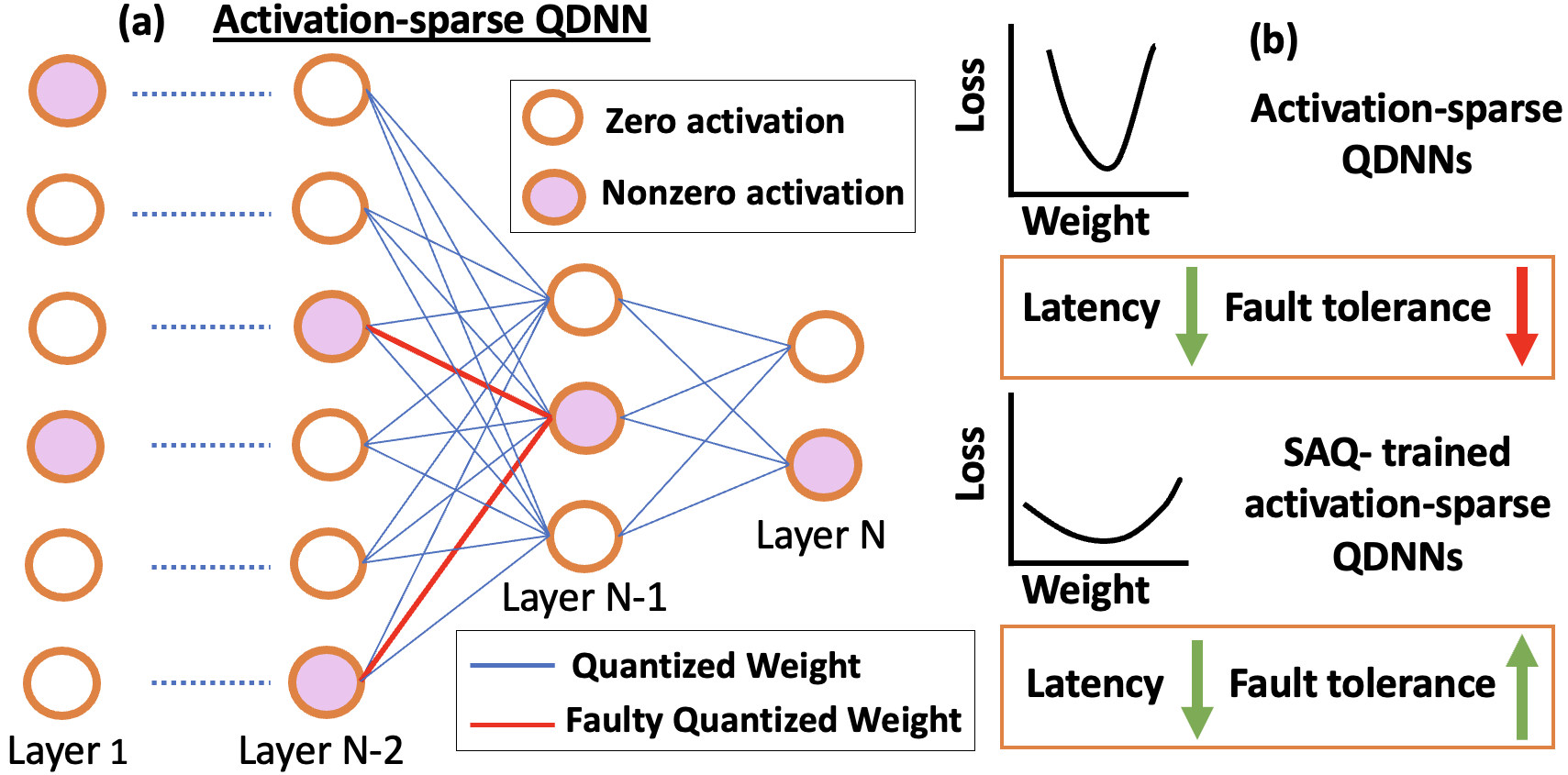}
  \caption{(a) Shows an activation-sparse QDNN (AS QDNN) with memory faults. We show that AS-QDNNs suffer larger accuracy degradation due to faults than their standard counterparts. (b) describes the trade-off between the latency benefits due to enhanced activation sparsity and reduced fault tolerance. To overcome this, we utilize sharpness-aware quantization (SAQ) training, which enhances the AS QDNNs fault tolerance by flattening its weight loss landscape.}
  \label{fig:overview}
\end{figure}

Another popular method to enhance DNN hardware efficiency is to sparsify the weights and activations. Various techniques to induce weight sparsity, also referred to as weight pruning, have led to significant resource savings with little or no accuracy loss \cite{pruning}. Recently, approaches to leverage and increase the activation sparsity have also gained attention to improve inference speed \cite{act_sparse1} \cite{act_sparse2}. Such techniques involving weight and activation sparsification can be used in conjunction with quantization to design QDNNs optimized for edge applications. 

While sparsified QDNNs offer promising attributes for resource-constrained systems, their deployment in real world settings also needs to consider hardware defects.  DNN  weights are generally stored on-chip in a memory which experiences various types of faults such as stuck-at and bit-flip faults. These faults corrupt the weight values and degrade system accuracy. Although the impact of faults in conventional DNNs is well understood, the fault tolerance of weight/activation sparse DNNs has been sparsely studied. Some works have explored the impact of faults and non-idealities on weight pruned models \cite{faults_non_ideal}\cite{faults_pruning}. However, the understanding of the impact of faults on \textit{activation-sparse} DNNs (AS QDNNs) is lacking. Moreover, there is a need for techniques that can alleviate the adverse effect of faults on the accuracy of AS QDNNs.

In this paper, we address these critical needs by (a) analyzing the inference accuracy of AS QDNNs in the presence of memory faults and (b) enhancing their fault tolerance by utilizing Sharpness-Aware Quantization (SAQ) \cite{saq}. The key contributions of our work are (summarized in Figure~\ref{fig:overview}) :
\begin{itemize}
    \item We show that although an increase in activation sparsity can lead to latency benefits, it comes at the expense of reduced fault tolerance and lower inference accuracy of QDNNs. To the best of our knowledge, this is the first work exploring the latency-accuracy trade-off for \textit{AS QDNNs} in the presence of faults.  
    \item  Using the weight loss landscape visualization method \cite{visualize}, we establish that the higher sensitivity of AS QDNNs to faults is attributed to sharper minima in their loss landscape (compared to the standard QDNNs).
    \item Based on the above finding, we employ SAQ to mitigate the impact of faults on system accuracy \cite{saq}. We show that the SAQ-training increases the inference accuracy of AS and standard QDNNs by up to 19.50\% and 15.82\% (compared to conventional training schemes) by reducing the sharpness of the weight loss landscape. 
    \item We also show that SAQ-trained AS QDNNs have higher inference accuracy than their standard counterparts trained without SAQ. This enables the design of QDNNs which are both activation-sparse and fault tolerant. 
\end{itemize}

\section{Background and Related Work}
\label{sec:background}

\subsection{Activation sparsity in DNNs}
Activation sparsity refers to the prevalence of a large number of zero-valued DNN activations. A common activation function in DNNs is the rectified linear unit (ReLU), which outputs a zero for every negative input, leading to a high activation sparsity. This motivates the design of sparsity-aware DNN hardware, that can skip the computations involving zero-valued activations, leading to a reduction in the inference latency \cite{act_sparse3}. Unlike weight sparsity, activation sparsity is dynamic in nature, which means that the number and location of the zero values vary from input to input. When utilized properly, the inherently large activation sparsity can be exploited by sparsity-aware DNN accelerators for sizeable latency improvements over standard accelerators  \cite{accelerator} \cite{act_sparse3}. 

Consequently, algorithmic techniques to \textit{enhance} the activation sparsity have gained interest in recent times. For example, the work in \cite{act_sparse2} adds the $L_{1}$ norm of the activations (\(||x_{l,n}||_{1}\)) to the original loss function to incentivize sparsity:
\begin{equation}
    L_{reg}(x,w) = L_{0}(x,w) + \sum^{N}_{n=1}\sum^{L}_{l=1}\alpha_{l}||x_{l,n}||_{1}
\label{eq1}
\end{equation}

Here, $L_{reg}(x,w)$ and $L_{0}(x,w)$ are the new and original loss functions, respectively, $L$ is the number of layers in the DNN, $N$ is the batch size and $\alpha_{l}$ is the regularization constant per layer. By penalizing the dense activations using $L1$ regularization, the work in \cite{act_sparse2} exhibits up to 60\% increase in sparsity with negligible loss in inference accuracy for image classification. The work in \cite{act_sparse1} uses a Hoyer sparsity metric based regularizer and a variant of ReLU to boost the activation sparsity of DNNs. It should be noted that these techniques are not equivalent to "pruning" the activations \cite{pruning}. While pruning permanently sets the parameters to zero for all inputs, the activation sparsification techniques do not remove any of the activations. Rather, they ensure that a low percentage of activations are non-zero for all inputs \textit{on an average}. 

In this paper, we use the $L1$ regularization based approach to train activation-sparse (AS) QDNNs. We will refer to the QDNNs trained without this regularization as standard QDNNs and the ones trained with it as AS QDNNs. 

\subsection{Memory faults in DNN accelerators}
Aggressive scaling and exploration of new memory technologies have heightened the importance of studying memory faults. Faults corrupt a percentage of the stored weights, leading to erroneous computations. Two common memory fault models are bit-flip faults and stuck-at (SA) faults. Stuck-at one (SA1) and stuck-at zero (SA0) faults fix the value of a bitcell \textit{unalterably} to  '1' and '0', respectively. Conversely, bit-flip faults invert the value in the bitcell (from '0' to '1' or vice versa). The SA fault model is generally used to describe permanent defects such as fabrication defects, whereas the bit-flip fault model is used to describe transient phenomena such as half-select read disturbance and alpha particle strikes.  

Understanding the impact of memory faults on DNN accuracy has attracted attention in recent times. Various works have analyzed the fault tolerance of both floating-point DNNs and QDNNs \cite{fault_survey}. Some works have also studied the accuracy of pruned networks in the presence of faults \cite{faults_pruning}, circuit non-idealities and variations \cite{faults_non_ideal}, showing a larger vulnerability of pruned DNNs to faults and other non-idealities than their unpruned counterparts. However, to the best of our knowledge, no work has analyzed the impact of faults on AS QDNNs. 

 Multiple hardware and algorithmic solutions have been analyzed for mitigating the impact of faults. From the hardware perspective, the work in \cite{hardware_mitigation1} proposes a selective duplication strategy to replicate the vulnerable portions of the DNNs to enhance fault tolerance. The work in \cite{tfix} leverages the natural redundancy present in compute-in-memory based ternary bitcells to reduce fault impact. From the algorithmic perspective, the work in \cite{software_mitigation1} uses error-correcting output codes to reduce DNN sensitivity to variations and SA faults. Works like \cite{bnn-flip} and \cite{mapping1} use weight transformations to reduce the amount of unmasked faults, leading to increased fault tolerance. Fault-aware training/finetuning is also a promising approach but is difficult to implement when the fault data is unavailable \cite{fat}. 

In this work, we extensively investigate the impact of SA and bit-flip faults on the performance of standard and AS QDNNs. Additionally, we introduce SAQ-based training as a strategy for fault tolerance. SAQ can be used alongside hardware techniques and offers protection against multiple types of faults, as discussed subsequently.

%
%
%

\subsection{Sharpness-Aware Quantization (SAQ)}
The sharpness of the minima of the loss landscape that a DNN converges to during training has a key impact on its generalization capability \cite{generalization1}. Both theoretically and empirically, it has been shown that convergence to a flat minima improves generalization. Hence, training schemes such as sharpness-aware minimization (SAM) are gaining attention  \cite{sam}. SAM simultaneously minimizes the loss value and the sharpness of the weight loss landscape to learn the optima which have uniformly low loss values in their neighbourhood. DNNs trained with SAM have displayed state-of-the-art accuracies for several datasets and models. However, SAM is not as effective in QDNNs as it does not account for weight quantization. Sharpness-aware quantization (SAQ) is a variant of SAM designed for QDNNs \cite{saq}. SAQ simultaneously minimizes the loss value and flattens the loss curvature near the minima by minimizing the loss function with adversarially perturbed quantized weights. This leads to better generalization and improved accuracy of QDNNs. In this work, we advance the application of SAQ to train fault-tolerant QDNNs, based on the intuition that QDNNs converged at flatter minima will be less sensitive to the weight perturbations due to faults.

\section{Impact of Faults on Activation-Sparse QDNNs}
\label{sec:impact}
\subsection{Experimental Framework}

To explore the latency benefits and the impact of faults on activation-sparse (AS) QDNNs, we use the following methodology. First, two 4-bit standard QDNNs (baselines), viz. LeNet-5 with the FashionMNIST (FMNIST) dataset and ResNet-18 with the CIFAR-10 dataset are trained using the quantization framework in \cite{google_quant}. Then, we train the AS QDNNs by combining the $L1$ activation regularization \cite{act_sparse2} with the quantization framework in \cite{google_quant}. We focus on these model architectures because of their compact size, making them suitable for resource-constrained edge accelerators.  The standard and AS QDNNs are trained to have nearly equal inference accuracies (with $\leq 0.5\%$ difference) for a fair comparison. 

After training, we evaluate their latencies on sparsity-aware hardware based on the work in \cite{act_sparse3}, to quantify the benefits of enhanced sparsity. Then, to study their fault tolerance, bit-flip faults and stuck-at faults (SA0 and SA1 individually) are randomly and uniformly injected into each model. Monte Carlo based fault injection experiments are performed, and the mean value of the inference accuracy is reported for different fault rates and types of faults (see Fig.~\ref{fig:total_saq}). We perform experiments for fault rates 0\% - 5\% and 0\% - 3\% for the LeNet-5 and ResNet-18 QDNNs, respectively. ResNet-18 QDNNs were analyzed up to a fault rate of 3\%, as higher rates caused severe accuracy drops, rendering the QDNNs unusable.

To better understand the differences in fault tolerance, we visualize the weight loss landscape for standard and AS QDNNs. Loss landscape visualization has been used in various works to enhance understanding of how well a model generalizes \cite{generalization1}. To our knowledge, this is the first work to examine the shape of the minima in relation to the QDNN's fault tolerance. We use the normalization-based visualization technique described in \cite{visualize} to generate the weight loss landscapes.

\subsection{Results}
\subsubsection{Latency benefits of Enhanced Activation Sparsity}
First, we analyse the activation sparsity increase due to $L1$ activation regularization. Figures~\ref{fig:sparsity}a and~\ref{fig:sparsity}b show the percentage of zero activations (averaged over all layers and test inputs) for LeNet-5 and ResNet-18 QDNNs, respectively. We observe an activation sparsity increase from 43\% (with standard training) to 84\% (with $L1$ activation regularization)  for LeNet-5 and from  53\% to 75\% for ResNet-18 QDNNs. We further analyze the sparsity of AS QDNNs in faulty settings. We observe that even at a 5\% (3\%) bit flip fault rate, the LeNet-5 (ResNet-18) AS QDNN shows only a $\sim$4\% ($\sim$2\%) reduction in sparsity. This signifies that the gain in activation sparsity is sustained even in the presence of faults.

\begin{figure}[t!]
\centering
  \includegraphics[width = 0.9\linewidth]{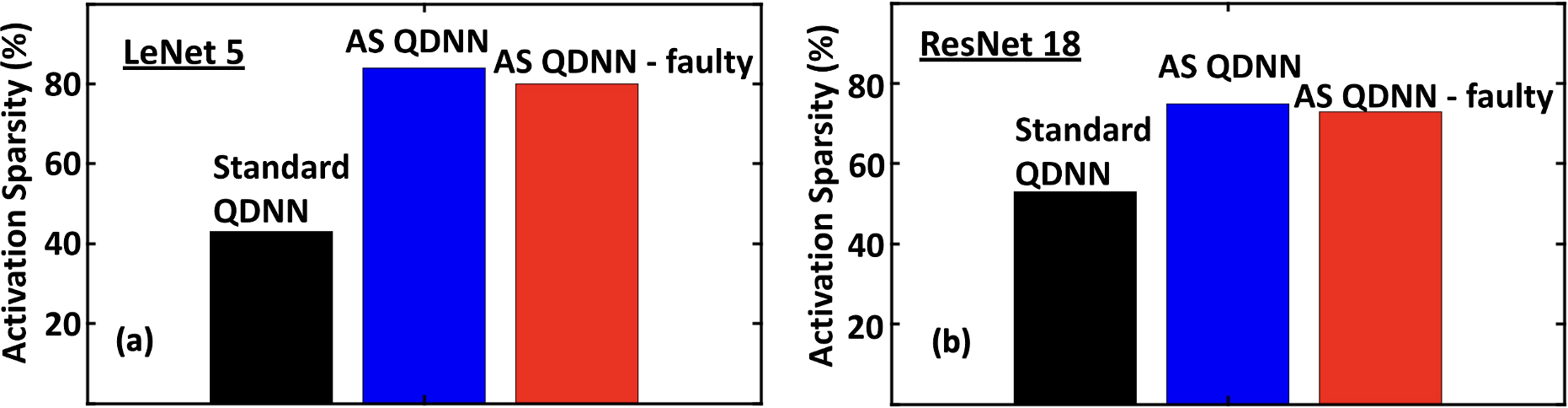}
  \caption{The activation sparsities of (a) LeNet 5 and (b) ResNet 18 standard QDNNs and activation sparse (AS) QDNNs in both fault-free and faulty settings. The activation sparsity of LeNet 5 and ResNet 18 AS QDNNs is 95.34\% and 41.51\% higher than their standard counterparts, and is sustained in faulty environments.}
  \label{fig:sparsity}
\end{figure}

\begin{figure}[b!]
\centering
  \includegraphics[width = 0.95\linewidth]{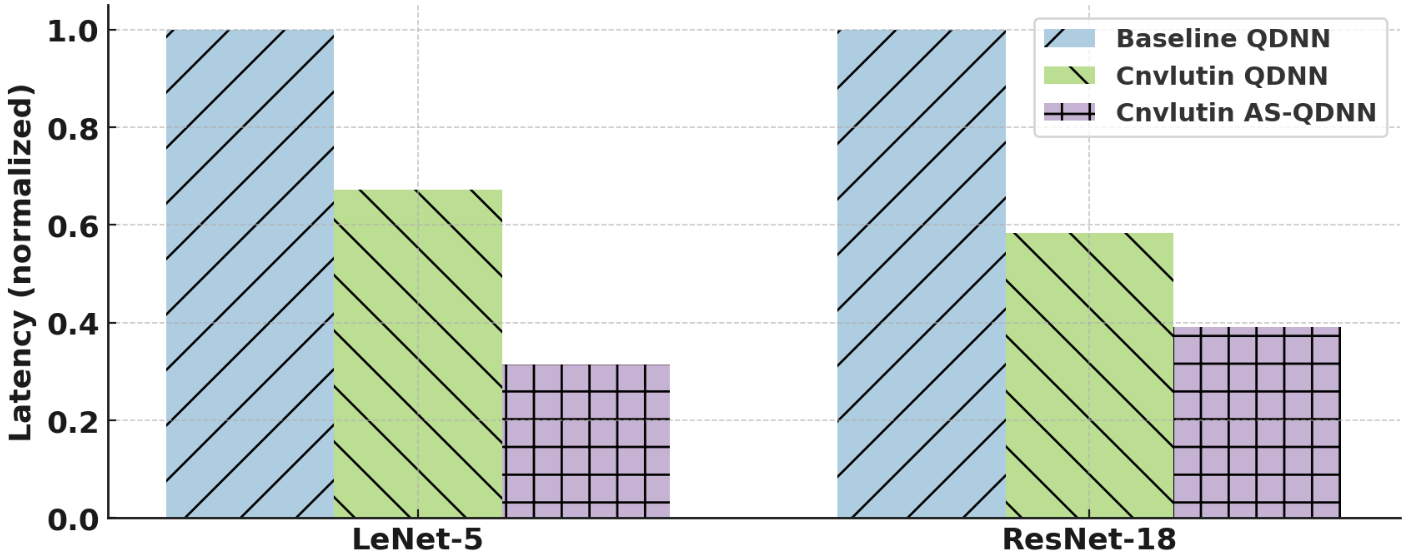}
  \caption{Latencies of various QDNNs deployed on DNN accelerators}
  \label{fig:latency_improvement}
\end{figure}

Next, we assess the latency benefits provided by the increased sparsity on a sparsity-aware accelerator based on the Cnvlutin architecture \cite{act_sparse3}. Cnvlutin skips computations with zero-valued activations, exploiting activation sparsity for speedup. For enabling zero-skipping, Cnvlutin incurs a mild area overhead of 4.49\% over the baseline (details in \cite{act_sparse3}). Additionally, as noted in \cite{act_sparse3}, Cnvlutin's power consumption is 7\% lower (on an average) than the baseline. Thus, Cnvlutin reduces latency without increasing power consumption.

Fig.~\ref{fig:latency_improvement} shows the normalized latencies of the ResNet-18 and LeNet-5 QDNNs. By exploiting the \textit{inherent} activation sparsity, Cnvlutin provides a 41.6\% and 32.9\% latency reduction over the baseline (without zero-skipping), for ResNet-18 and LeNet-5 standard QDNNs, respectively. Further, we observe that the ResNet-18 and LeNet-5 AS-QDNNs exhibit 60.9\% and 68.7\% lower latency when deployed on the Cnvlutin architecture than the standard QDNNs on the baseline architecture. Importantly, the increase in the sparsity from 53\% (43\%) to 75\% (84\%) when going from ResNet-18 (LeNet-5) QDNN to AS-QDNN leads to a latency reduction of 32.9\% (53.3\%). Thus,sparsity enhancement directly leads to latency reduction.

\subsubsection{Impact of faults on Inference Accuracy}
Next, we compare the fault tolerance of AS QDNNs and standard QDNNs, to gauge whether the sparsity enhancement has any effect on the accuracies of the models in the presence of faults. The dashed lines in Figures~\ref{fig:saq_lenet_bit_flip}-~\ref{fig:saq_resnet_sa1} show the impact of bit-flip, SA0 and SA1 faults on standard and AS LeNet-5 and ResNet-18 QDNNs. We observe that the AS QDNNs have lower inference accuracy in the presence of faults compared to the standard QDNNs. For bit-flip faults, the accuracy of LeNet-5 and ResNet-18 AS QDNNs is 2.40\%-11.13\% and 0.52\%-8.00\% lower (absolute difference in accuracy) than their equivalent standard QDNNs, respectively. This trend holds for SA0 and SA1 faults, albeit with lower accuracy degradation than bit-flip faults as the SA faults can potentially be masked \cite{bnn-flip}. These results signify that both AS QDNNs are \textit{less} fault tolerant than their standard QDNN counterparts. This implies that the latency benefits associated with increased activation sparsity comes at the price of higher vulnerability to faults. 

\begin{figure}[t!]
\centering
  \includegraphics[width = 0.88\linewidth]{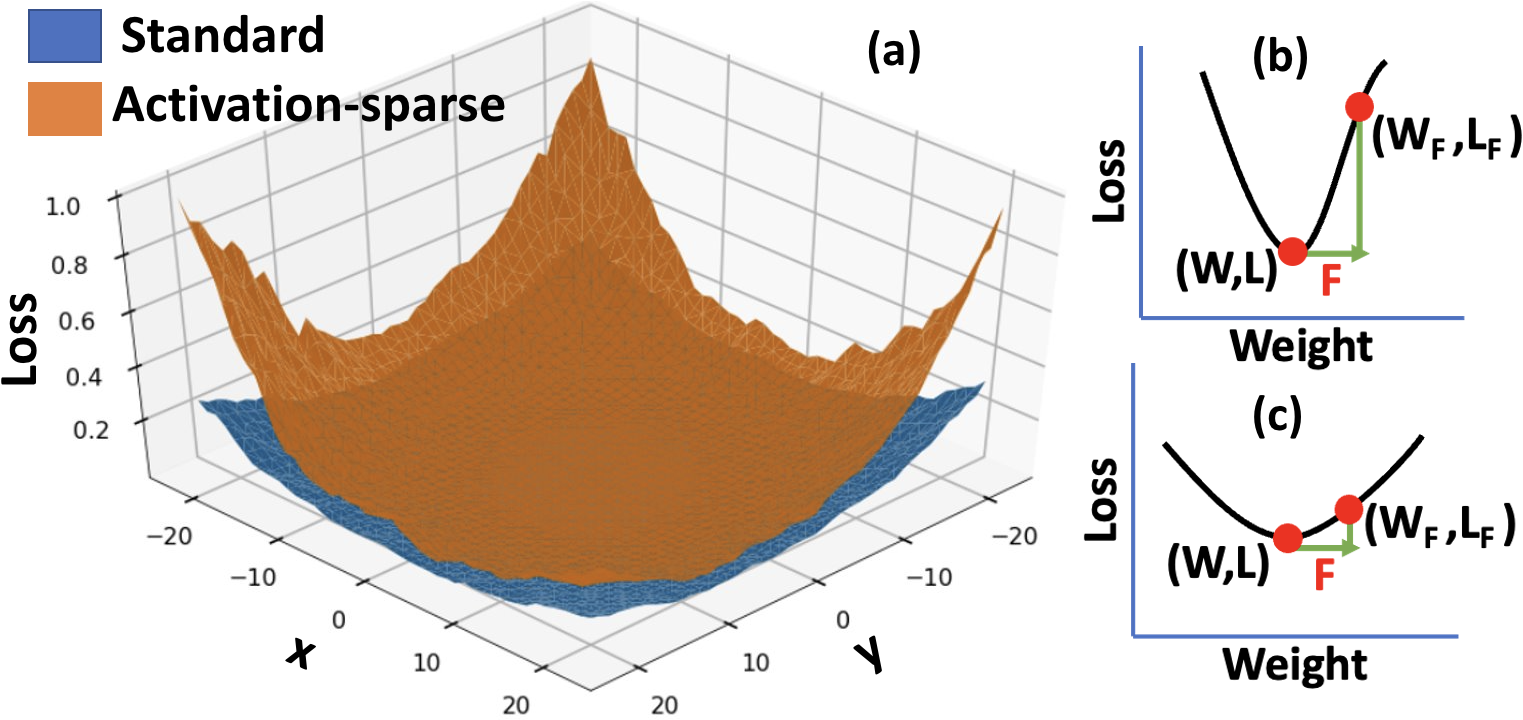}
  \caption{(a) The weight loss landscape of the standard and AS LeNet-5 QDNNs visualized using the technique in \cite{visualize}. \textit{x} and \textit{y} are normalized random directions. (b) and (c) show the impact of a fault (F) on the loss value in a landscape with (b) sharp and (c) flat minima. The fault causes a larger change in the loss value in the former.}
  \label{fig:sharp}
\end{figure}

\begin{figure*}[t!]
\centering
\captionsetup{justification=centering}
\begin{subfigure}{0.31\textwidth}
  \includegraphics[width=\linewidth]{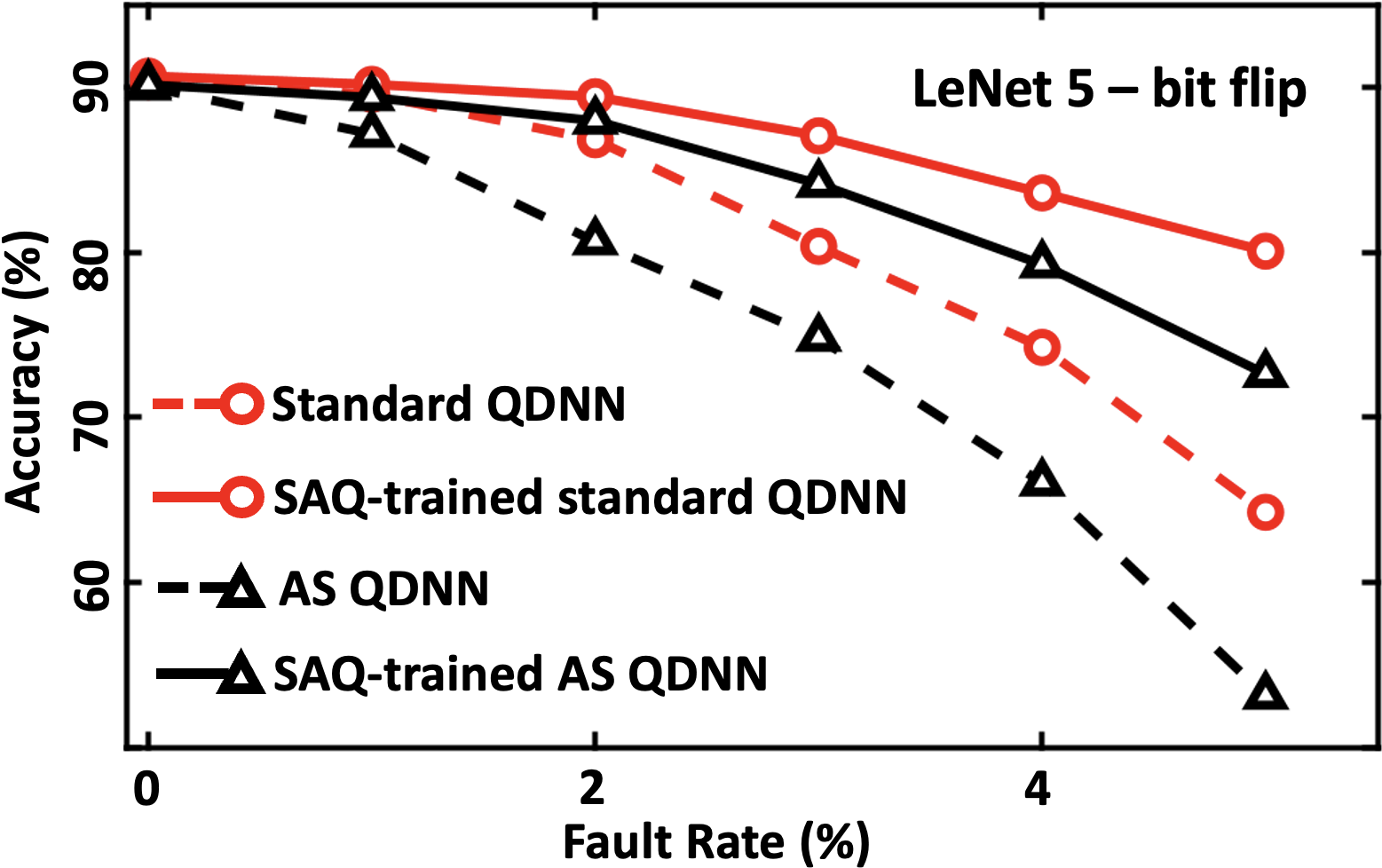}
  \caption{Bit-flip faults on LeNet-5 QDNNs}
  \label{fig:saq_lenet_bit_flip}
\end{subfigure}%
~
\begin{subfigure}{0.31\textwidth}
  \includegraphics[width=\linewidth]{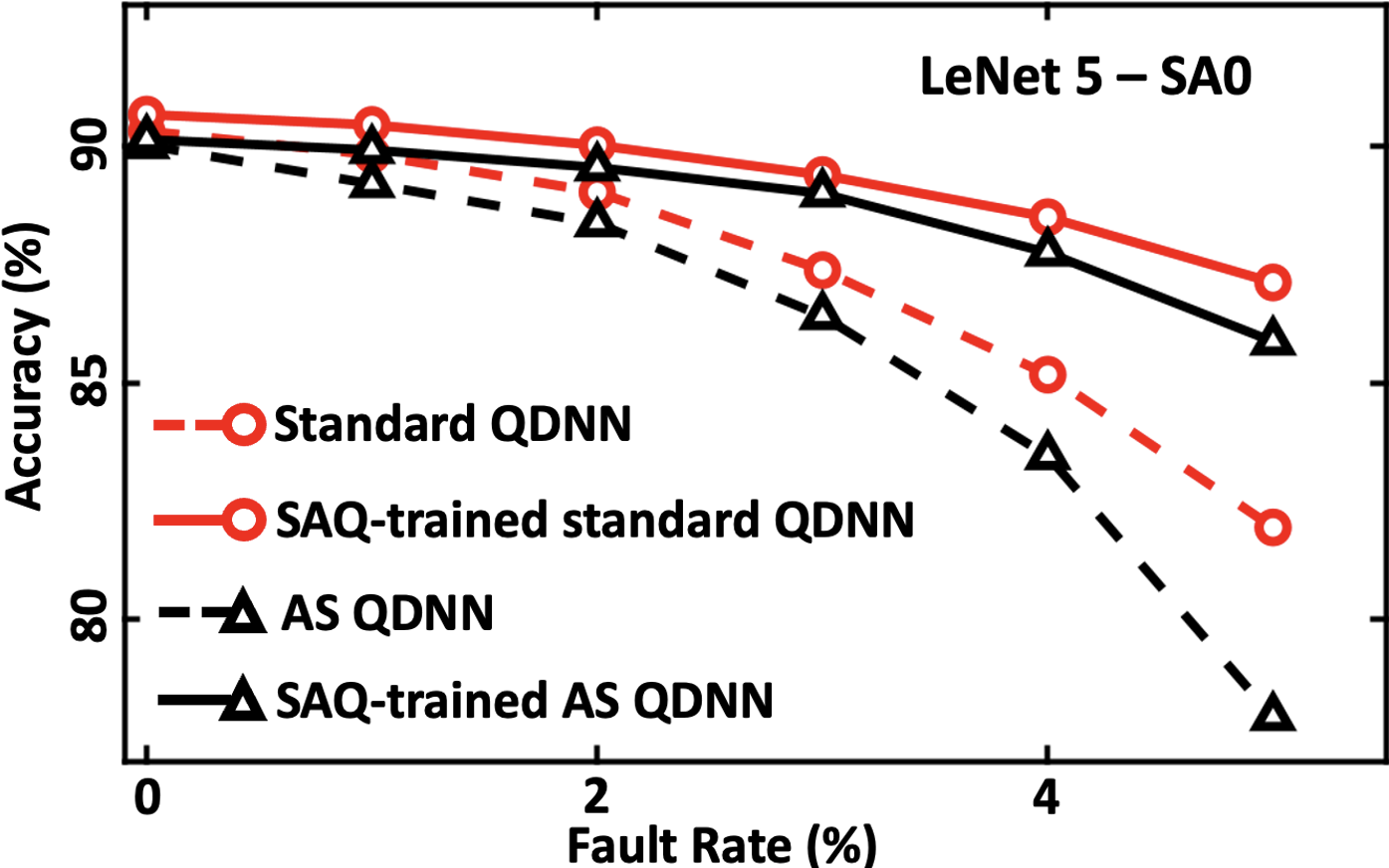}
  \caption{SA0 faults on LeNet-5 QDNNs}
  \label{fig:saq_lenet_sa0}
\end{subfigure}%
~
\begin{subfigure}{0.31\textwidth}
  \includegraphics[width=\linewidth]{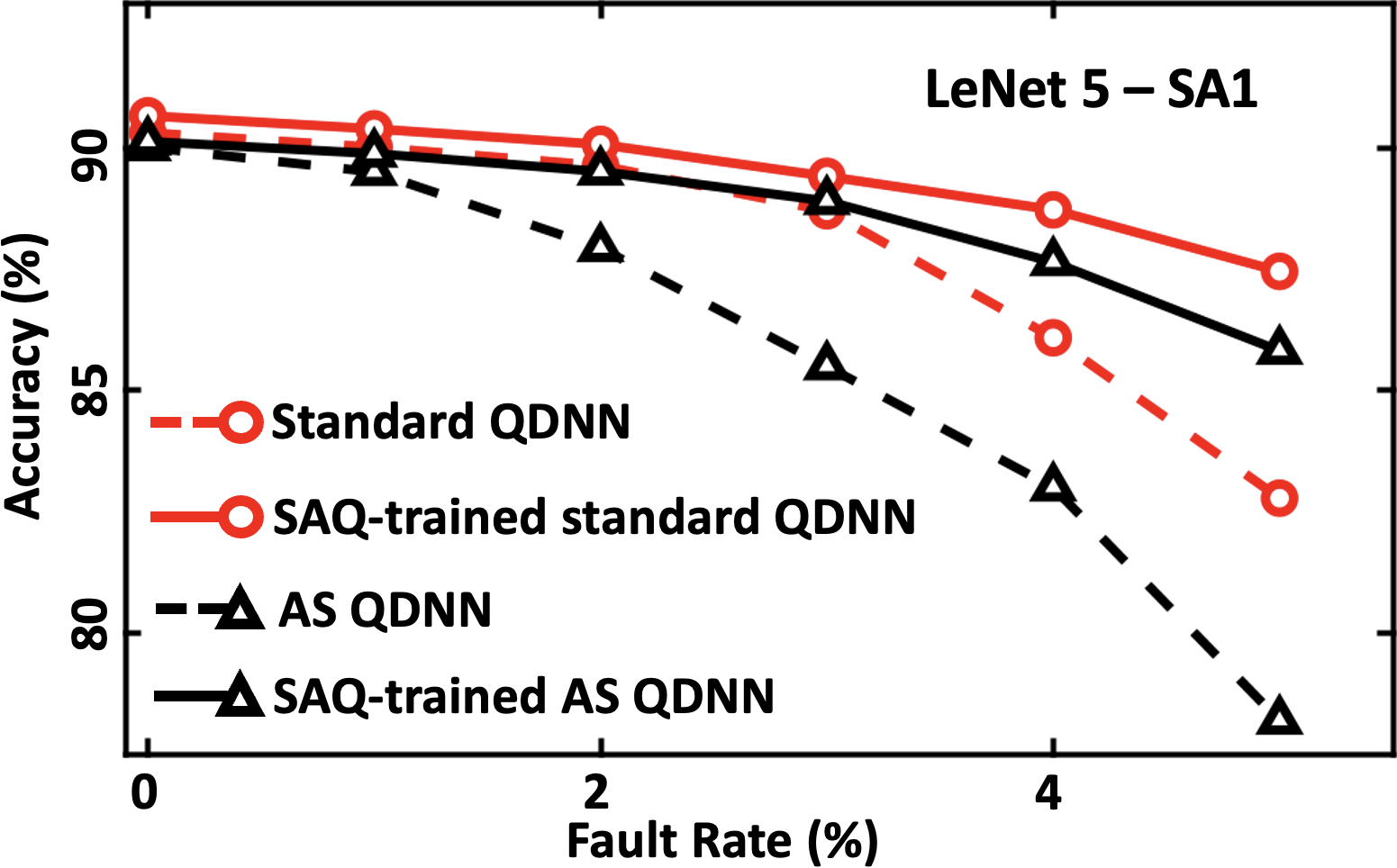}
  \caption{SA1 faults on LeNet-5 QDNNs}
  \label{fig:saq_lenet_sa1}
\end{subfigure}%
~
\vspace{4mm}

\begin{subfigure}{0.31\textwidth}
  \includegraphics[width=\linewidth]{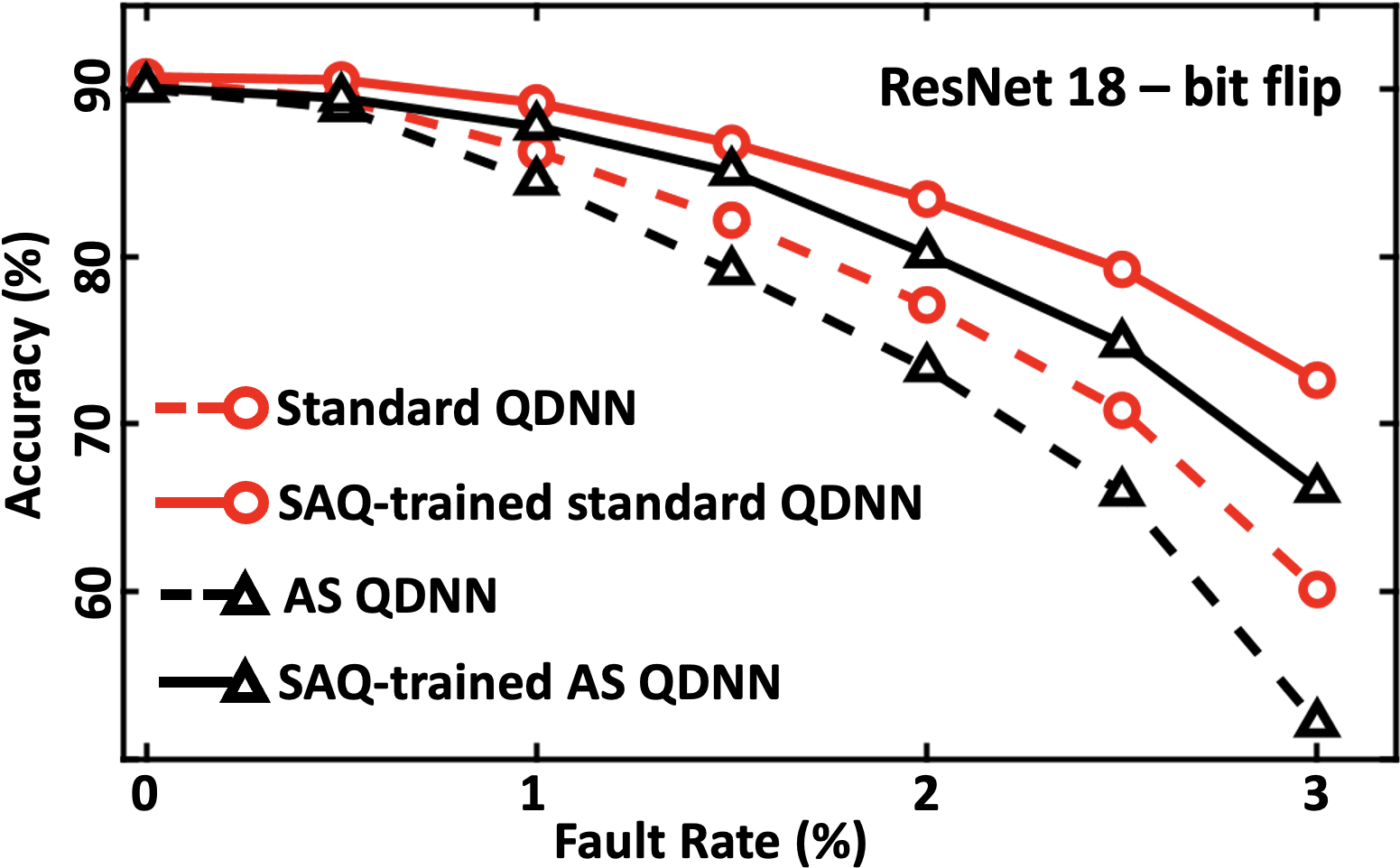}
  \caption{Bit-flip faults on ResNet-18 QDNNs}
  \label{fig:saq_resnet_bit_flip}
\end{subfigure}%
~
\begin{subfigure}{0.31\textwidth}
  \includegraphics[width=\linewidth]{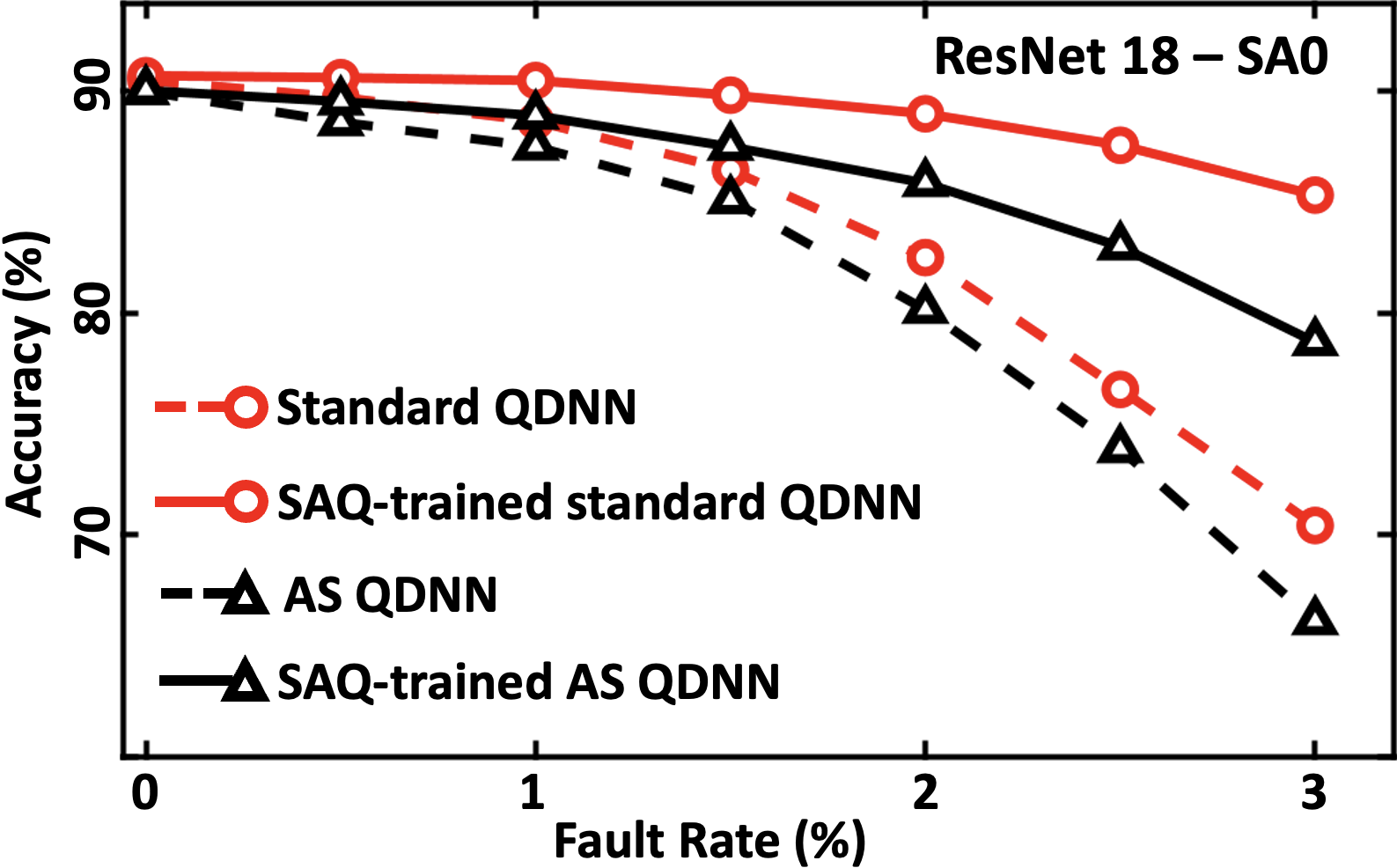}
  \caption{SA0 faults on ResNet-18 QDNNs}
  \label{fig:saq_resnet_sa0}
\end{subfigure}%
~
\begin{subfigure}{0.31\textwidth}
  \includegraphics[width=\linewidth]{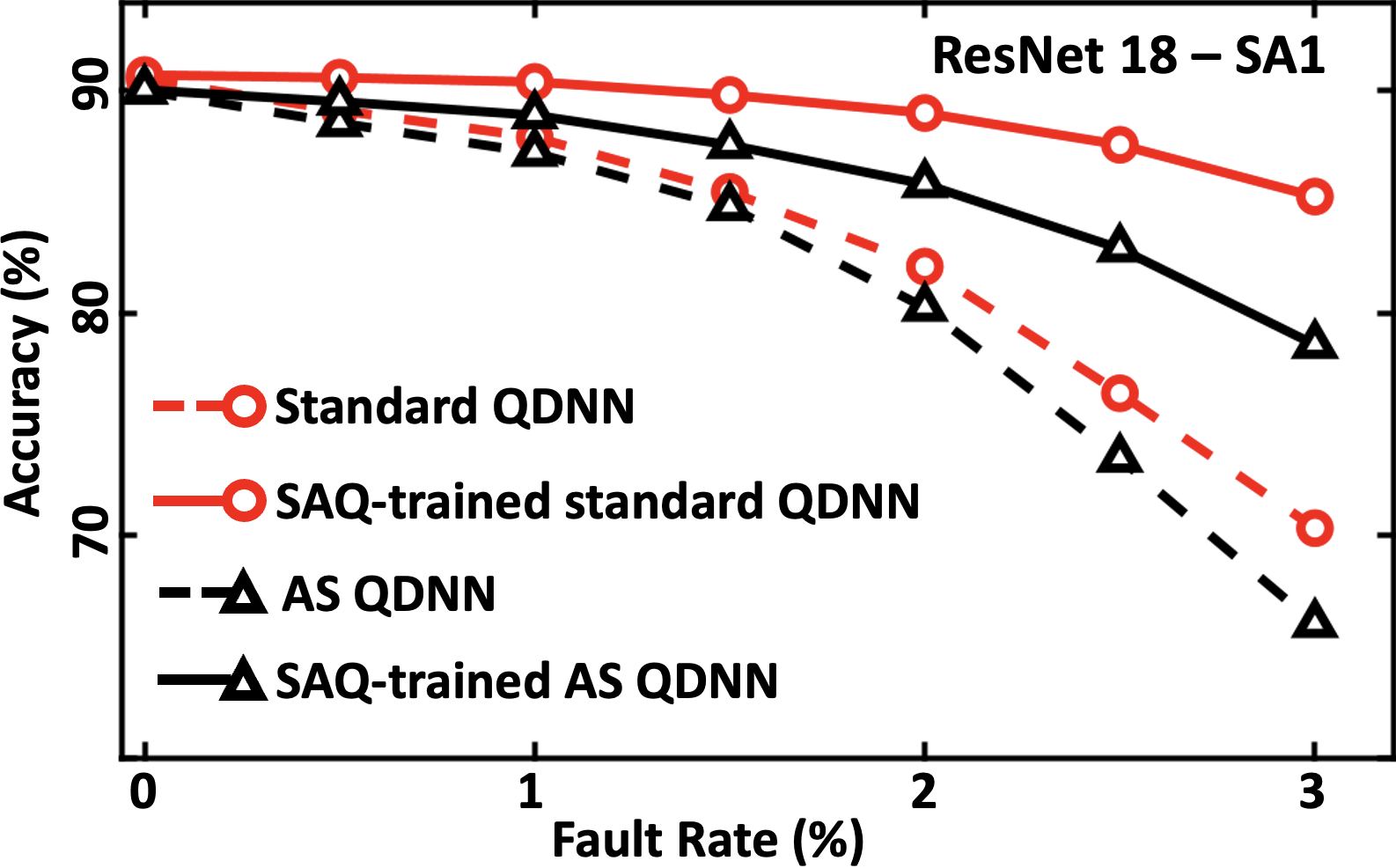}
  \caption{SA1 faults on ResNet-18 QDNNs}
  \label{fig:saq_resnet_sa1}
\end{subfigure}%
~
\vspace{4mm}
\caption{Comparison of the impact on classification accuracy for different fault scenarios for both SAQ trained and conventionally trained standard and activation-sparse (AS) QDNNs. It can be seen that AS QDNNs have lesser fault tolerance than their standard counterparts. Also, SAQ-trained QDNNs display higher fault tolerance than their conventionally trained equivalents. }
\label{fig:total_saq}

\end{figure*}
\subsubsection{Weight Loss Landscape visualization}

To explain the reduced fault tolerance of AS QDNNs, we visualize their weight loss landscapes. Figure~\ref{fig:sharp}a shows the weight loss landscape of the standard and AS LeNet-5 QDNN. Intuitively, a QDNN with a sharper minima would incur a larger change in the loss function value due to weight perturbation, as shown in Figure \ref{fig:sharp}b and c. We observe that the AS QDNN converges at a sharper minima, which explains its experimentally observed degraded fault tolerance. Thus, the increased activation sparsity due to the $L1$ activation regularization is accompanied by an increase in the sharpness of the minima and reduced fault tolerance. A similar trend was observed for ResNet-18.

To sum up, enhanced activation sparsity reduces the latency of sparsity-aware DNN accelerators, but degrades fault tolerance due to a sharper minima in the loss landscape. Based on this observation, we propose the utilization of sharpness-aware quantization (SAQ) based training to enhance the fault tolerance of AS QDNNs and show its effectiveness next. 

\section{Mitigation Strategy: SAQ}
\label{sec:mitigation}

\subsection{SAQ-based fault mitigation strategy}
SAQ flattens the weight loss landscape of the QDNNs, which motivates us to employ it for making the QDNN less sensitive to faults. To evaluate this approach, we utilize SAQ to train both standard and AS QDNNs. 

SAQ concurrently minimizes the loss value and loss sharpness by solving the following min-max optimization problem:
\begin{multline}
   \underset{w}{\min} \underset{||\epsilon||_{2} \leq \rho}{\max}((L_{S}(Q_{w}(w,b)) + \epsilon) - L_{S}(Q_{w}(w,b))) \\
   + L_{S}(Q_{w}(w,b)) + \frac{\lambda}{2}||w||^{2}_{2} 
\label{eq2}
\end{multline}

The first term in the equation defines the sharpness metric, which is the maximum change in the loss value for some weight perturbation $\epsilon$. The second term is the loss function itself and the third term is the standard $L2$ regularization term. Also, the perturbation $\epsilon$ is an adversarial one and is chosen such that the maximum sharpness is minimized. $\epsilon$ is given by the following equation and can have a maximum value of $\rho$.
\begin{equation}
\epsilon \simeq \hat{\epsilon} = \rho\frac{\nabla_{Q_{w}(w,b)}L_{S}(Q_{w}(w,b))}{||\nabla_{Q_{w}(w,b)}L_{S}(Q_{w}(w,b))||_{2}}   
\label{eq3}
\end{equation}

Here, $\nabla_{Q_{w}(w,b)}L_{S}(Q_{w}(w,b))$ is the gradient of the loss function with respect to the quantized weights. $\epsilon$ is estimated using a forward and backward pass through the QDNN. Thus, compared to conventional training, SAQ training requires double the number of forward/backward passes per epoch. For a fair comparison of SAQ-based QDNNs with conventionally trained ones, we train SAQ models for half as many epochs. We select the $\rho$ hyperparameter to maximize the fault-free accuracy (fault rate of 0\%) for the QDNNs. Using this framework, we analyse the effectiveness of SAQ in increasing the fault tolerance for both standard QDNNs and AS QDNNs.

\subsection{Results}
 First, we compare standard QDNNs trained conventionally and with SAQ. Figure~\ref{fig:total_saq} shows that SAQ-trained standard QDNNs exhibit larger fault tolerance than their conventionally trained counterparts. For example, for a bit-flip fault rate of 3\%, SAQ-training increases the inference accuracy of LeNet-5 and ResNet-18 by 6.70\% and 12.49\%, respectively. 
 
 Next, we compare the accuracies of SAQ-trained and conventionally trained AS QDNNs. The $L1$ regularization constant is kept the same for both SAQ-trained and conventionally trained AS QDNNs, ensuring similar activation sparsity. The trends observed for AS QDNNs are similar to those for standard QDNNs. For a bit-flip fault rate of 3\%, SAQ training enhances the inference accuracy of LeNet-5 and ResNet-18 AS QDNNs by 9.34\% and 13.96\%. 

An interesting observation here is that SAQ-trained AS QDNNs exhibit a higher accuracy in the presence of faults than standard QDNNs without SAQ. For example, our experiments show that the SAQ-trained LeNet 5 AS QDNN has a 8.37\% higher accuracy than the conventionally trained LeNet 5 standard QDNN at fault rate = 5\%. Thus, SAQ-trained AS QDNNs have both the benefits of superior fault tolerance \textit{and} lower latency (compared to conventionally trained standard QDNNs), making them highly suitable for edge applications. Thus,
training with SAQ mitigates the adverse impact that activation sparsity augmentation has on the fault tolerance.


\section{Conclusion}
\label{sec:conclusion}

In this work, we show that the increase in activation sparsity (and the associated speedup) comes at the price of degraded inference accuracy in the presence of faults, with activation sparse (AS) QDNNs showing up to 11.13\% lower accuracy than standard QDNNs. To explain this reduced fault tolerance, we visualize the weight loss landscapes for standard and AS QDNNs and establish that AS QDNNs exhibit a sharper minima leading to lower fault tolerance. Based on this observation, we employ sharpness-aware quantization (SAQ) based training, which flattens the loss landscape and enhances fault tolerance. Our results show that AS and standard QDNNs trained with SAQ have up to 19.50\% and 15.82\% higher accuracy than their conventionally trained counterparts. We also observe that SAQ-trained AS QDNNs have higher inference accuracy than conventionally trained standard QDNNs, enabling QDNNs which are not only activation-sparse (and thus, low-latency) but also fault-tolerant.


\bibliography{references.bib}

\begin{thebibliography}{10}
\providecommand{\url}[1]{#1}
\csname url@samestyle\endcsname
\providecommand{\newblock}{\relax}
\providecommand{\bibinfo}[2]{#2}
\providecommand{\BIBentrySTDinterwordspacing}{\spaceskip=0pt\relax}
\providecommand{\BIBentryALTinterwordstretchfactor}{4}
\providecommand{\BIBentryALTinterwordspacing}{\spaceskip=\fontdimen2\font plus
\BIBentryALTinterwordstretchfactor\fontdimen3\font minus
  \fontdimen4\font\relax}
\providecommand{\BIBforeignlanguage}[2]{{%
\expandafter\ifx\csname l@#1\endcsname\relax
\typeout{** WARNING: IEEEtran.bst: No hyphenation pattern has been}%
\typeout{** loaded for the language `#1'. Using the pattern for}%
\typeout{** the default language instead.}%
\else
\language=\csname l@#1\endcsname
\fi
#2}}
\providecommand{\BIBdecl}{\relax}
\BIBdecl

\bibitem{accelerator}
W.~Li \emph{et~al.}, ``A precision-scalable deep neural network accelerator
  with activation sparsity exploitation,'' \emph{IEEE Trans. Comp.-Aid. Des.
  Integ. Circ. Sys.}, 2024.

\bibitem{google_quant}
B.~Jacob \emph{et~al.}, ``Quantization and training of neural networks for
  efficient integer-arithmetic-only inference,'' in \emph{2018 IEEE/CVF Conf.
  Comp. Vis. Pat. Rec.}, 2018, pp. 2704--2713.

\bibitem{pruning}
S.~Vadera and S.~Ameen, ``Methods for pruning deep neural networks,''
  \emph{IEEE Access}, vol.~10, pp. 63\,280--63\,300, 2022.

\bibitem{act_sparse1}
M.~Kurtz \emph{et~al.}, ``Inducing and exploiting activation sparsity for fast
  inference on deep neural networks,'' in \emph{Proc. 37th Int. Conf. Mach.
  Learn.}, 2020, pp. 5533--5543.

\bibitem{act_sparse2}
G.~Georgiadis, ``Accelerating convolutional neural networks via activation map
  compression,'' in \emph{2019 IEEE/CVF Conf. Comp. Vis. Pat. Rec. (CVPR)},
  2019, pp. 7078--7088.

\bibitem{faults_non_ideal}
A.~Bhattacharjee \emph{et~al.}, ``Examining and mitigating the impact of
  crossbar non-idealities for accurate implementation of sparse deep neural
  networks,'' in \emph{Proc. 2022 Conf. \& Exhib. Des., Auto. \& Test Eur.},
  2022, p. 1119–1122.

\bibitem{faults_pruning}
G.~Yuan \emph{et~al.}, ``Improving dnn fault tolerance using weight pruning and
  differential crossbar mapping for reram-based edge ai,'' in \emph{2021 22nd
  Int. Symp. Qual. Elec. Des. (ISQED)}, 2021, pp. 135--141.

\bibitem{saq}
J.~Liu \emph{et~al.}, ``Sharpness-aware quantization for deep neural
  networks,'' 2023.

\bibitem{visualize}
H.~Li \emph{et~al.}, ``Visualizing the loss landscape of neural nets,'' in
  \emph{Adv. Neur. Inf. Proces. Sys.}, vol.~31, 2018.

\bibitem{act_sparse3}
J.~Albericio \emph{et~al.}, ``Cnvlutin: Ineffectual-neuron-free deep neural
  network computing,'' in \emph{2016 ACM/IEEE 43rd Ann. Int. Symp. Comp. Arch.
  (ISCA)}, 2016, pp. 1--13.

\bibitem{fault_survey}
S.~Mittal, ``A survey on modeling and improving reliability of dnn algorithms
  and accelerators,'' \emph{Jour. Sys. Arch.}, vol. 104, p. 101689, 2020.

\bibitem{hardware_mitigation1}
F.~Libano \emph{et~al.}, ``Selective hardening for neural networks in fpgas,''
  \emph{IEEE Trans. Nuc. Sci.}, vol.~66, no.~1, pp. 216--222, 2019.

\bibitem{tfix}
A.~Malhotra \emph{et~al.}, ``Tfix: Exploiting the natural redundancy of ternary
  neural networks for fault tolerant in-memory vector matrix multiplication,''
  in \emph{2023 60th ACM/IEEE Des. Aut. Conf. (DAC)}, 2023, pp. 1--6.

\bibitem{software_mitigation1}
T.~Liu \emph{et~al.}, ``A fault-tolerant neural network architecture,'' in
  \emph{2019 56th ACM/IEEE Des. Aut. Conf. (DAC)}, 2019, pp. 1--6.

\bibitem{bnn-flip}
A.~Malhotra \emph{et~al.}, ``Bnn-flip: Enhancing the fault tolerance and
  security of compute-in-memory enabled binary neural network accelerators,''
  in \emph{2024 29th Asia S. Pac. Des. Aut. Conf. (ASP-DAC)}, 2024, pp.
  146--152.

\bibitem{mapping1}
B.~Zhang \emph{et~al.}, ``Handling stuck-at-fault defects using matrix
  transformation for robust inference of dnns,'' \emph{IEEE Trans. Comp.-Aid.
  Des. Integ. Circ. Sys.}, vol.~39, no.~10, pp. 2448--2460, 2020.

\bibitem{fat}
U.~Zahid \emph{et~al.}, ``Fat: Training neural networks for reliable inference
  under hardware faults,'' in \emph{2020 IEEE Int. Test Conf}, 2020, pp. 1--10.

\bibitem{generalization1}
Y.~Jiang \emph{et~al.}, ``Fantastic generalization measures and where to find
  them,'' in \emph{Int. Conf. Learn. Repres.}, 2020.

\bibitem{sam}
P.~Foret \emph{et~al.}, ``Sharpness-aware minimization for efficiently
  improving generalization,'' in \emph{Int. Conf. Learn. Repres.}, 2021.

\end{thebibliography}
\bibliographystyle{IEEEtran}

\end{document}